\documentclass[PP]{AEA}

\usepackage[abbr]{harvard}

\draftSpacing{1.5}
\newtheorem{algorithm}{Algorithm}
\begin{document}

\title{$L_2$Boosting for Economic Applications}
\author{Ye Luo and Martin Spindler\thanks{Luo: University of Florida, yeluo@ufl.edu.
 Spindler:  University of Hamburg, Hamburg Business School, Hamburg Center for Health Economics (hche) and Max Planck Society, martin.spindler@uni-hamburg.de. We thank participants and the discussant Hai Wang at the session Machine Learning in Econometrics at the AEA annual meeting 2017 in Chicago for valuable comments.}}
\pubMonth{Month}
\date{\today}
\pubYear{Year}
\pubVolume{Vol}
\pubIssue{Issue}
\maketitle

In the recent years more and more high-dimensional data sets, where the number of parameters $p$ is high compared to the number of observations $n$ or even larger, are available for applied researchers. Boosting algorithms represent one of the major advances in machine learning and statistics in recent years and are suitable for the analysis of such data sets. While Lasso has been applied very successfully for high-dimensional data sets in Economics, boosting has been underutilized in this field, although it has been proven very powerful in fields like Biostatistics and Pattern Recognition. We attribute this to missing theoretical results for boosting. The goal of this paper is to fill this gap and show that boosting is a competitive method for inference of a treatment effect or instrumental variable (IV) estimation in a high-dimensional setting. First, we present the $L_2$Boosting with componentwise least squares algorithm and variants which are tailored for regression problems which are the workhorse for most Econometric problems. Then we show how $L_2$Boosting can be used for estimation of treatment effects and IV estimation. We highlight the methods and illustrate them with simulations and empirical examples. For further results and technical details we refer to \cite{SL_2016} and \cite{SL_2017} and to the online supplement of the paper.

\section{$L_2$Boosting}
To define the boosting algorithm for linear models, we consider the following regression setting:
\begin{align} y_i=x_i'\beta+\varepsilon_i, \quad i=1,\ldots,n, \end{align}
 with vector $x_i=(x_{i,1},\ldots,x_{i,p_n})$ consisting of $p_n$ predictor variables, $\beta$ a $p_n$-dimensional coefficient vector, and a random, mean-zero error term $\varepsilon_i$, $\textrm{E}[\varepsilon_i|x_i]=0$.

We allow the dimension of the predictors $p_n$ to grow with the sample size $n$, and even be larger than the sample size, i.e., $dim(\beta)=p_n\gg n$. But we will impose a sparsity condition. This means that there is a large set of potential variables, but the number of variables which have non-zero coefficients, denoted by $s$, is small compared to the sample size, i.e.~ $s \ll n$.

$X$ denotes the $n \times p$ design matrix where the single observations $x_i$ form the rows. $X_j$ denotes the $j$th column of the design matrix, and $x_{i,j}$ the $j$th component of the vector $x_i$.
We consider a fixed design for the regressors. We assume that the regressors are standardized with mean zero and variance one, i.e., $\textrm{E}_n[x_{i,j}]=0$ and $\textrm{E}_n[x_{i,j}^2]=1$ for $j=1,\ldots,p$,

The basic principle of Boosting can be described as follows. We follow the interpretation of \cite{nr:breiman:1998} and \cite{nr:friedman:2001} of Boosting as a functional gradient descent optimization (minimization) method. The goal is to minimize a loss function, e.g., an $L_2$-loss or the negative log-likelihood function of a model, by an iterative optimization scheme. In each step the (negative) gradient which is used to update the current solution is modeled and estimated by a parametric or nonparametric statistical model, the so-called base learner. The fitted gradient is used for updating the solution of the optimization problem. A strength of boosting, besides the fact that it can be used for different loss functions, is its flexibility with regard to the base learners. We then repeat this procedure until some stopping criterion is met. The act of stopping is crucial for boosting algorithms, as stopping too late or never stopping leads to overfitting and therefore some kind of penalization is required. A suitable solution is to stop early, i.e., before overfitting takes place. \textquotedblleft Early stopping\textquotedblright\ can be interpreted as a form of penalization. Similar to LASSO, early stopping might induce a bias through shrinkage.

The literature has developed many different forms of boosting algorithms. In this paper we consider $L_2$Boosting with componentwise linear least squares, as well as two variants. All three are designed for regression analysis. \textquotedblleft $L_2$\textquotedblright refers to the loss function, which is the typical sum-of-squares of the residuals $Q_n(\beta)=\sum_{i=1}^n (y_i - x_i' \beta)^2$ typical in regression analysis. In this case, the gradient equals the residuals. \textquotedblleft Componentwise linear least squares\textquotedblright refers to the base learners. We fit the gradient (i.e. residuals) against each regressor ($p$ univariate regressions) and select the predictor/variable which correlates most highly with the gradient/residual, i.e., decreases the loss function most, and then update the estimator in this direction. We next update the residuals and repeat the procedure until some stopping criterion is met. We consider $L_2$Boosting and two modifications: the \textquotedblleft classical\textquotedblright one which was introduced in \cite{nr:friedman:2001} and refined in \cite{nr:buhlmann.yu:2003} for regression analysis, an orthogonal variant and post-$L_2$Boosting. In signal processing and approximation theory, the first two methods are known as the pure greedy algorithm (PGA) and the orthogonal greedy algorithm (OGA) in the deterministic setting, i.e. in a setting without stochastic error terms.

\begin{algorithm}[$L_2$Boosting]
\begin{enumerate}
    \item Start / Initialization: $\beta^0 = 0$ ($p$-dimensional vector), $f^0=0$, set maximum number of iterations $m_{stop}$ and set iteration index $m$ to $0$.
    \item At the $(m+1)^{th}$ step, calculate the residuals $U_i^m=y_i - x_i' \beta^m$.
    \item For each predictor variable $j=1,\ldots,p$ calculate the correlation with the residuals:
    \begin{align*} \gamma^m_{j}:=\frac{\sum_{i=1}^n U_i^m x_{i,j}}{\sum_{i=1}^n x_{i,j}^2}=\frac{<U^m,x_j>_n}{\textrm{E}_n[x_{i,j}^2]}.  \end{align*}
    Select the variable $j^m$ that is the most correlated with the residuals, i.e., $\max_{1\leq j\leq p}|corr(U^{m},x_{j})|$\footnote{Equivalently, which fits the gradient best in a $L_2$-sense.}.
    \item Update the estimator: $\beta^{m+1}:=\beta^{m}+\gamma^m_{j^m} e_{j^m}$ where $e_{j^m}$ is the $j^m$th index vector
and
    $f^{m+1}:=f^{m}+\gamma^m_{j^m}x_{j^m}$
 \item Increase $m$ by one. If $m<m_{stop}$, continue with (2); otherwise stop.
\end{enumerate}
\end{algorithm}

Moreover, we consider two variants, namely post-$L_2$Boosting (post-BA) and orthogonal $L_2$Boosting (oBA).

Post-$L_2$Boosting is a post-model selection estimator that applies ordinary least squares (OLS) to the model selected by the first-step, namely $L_2$Boosting. To define this estimator formally, we make the following definitions: $T:=supp(\beta)$ and $\hat{T}:=supp(\beta^{m^*})$, the support of the true model and the support of the model estimated by $L_2$Boosting as described above with stopping at $m^*$. A superscript $C$ denotes the complement of the set with regard to $\{1,\ldots,p\}$. In the context of LASSO, OLS after model selection was analyzed in \cite{belloni:2013}.
Given the above definitions, the post-model selection estimator or OLS post-$L_2$Boosting estimator will take the form
\begin{align}
\tilde{\beta}= arg min_{\beta \in \textrm{R}^p} Q_n(\beta): \beta_j=0 \forall j \in \hat{T}^C,
\end{align}

where $Q_n(\beta)=\sum_{i=1}^n (y_i - x_i' \beta)^2$.
For oBA, only the updating step is changed: an orthogonal projection of the response variable is conducted on all the variables which have been selected up to this point. The advantage of this method is that any variable is selected at most once in this procedure, while in the previous version the same variable might be selected at different steps which makes the analysis far more complicated.
More formally, the method can be described as follows by modifying Step (4):
\begin{algorithm}[Orthogonal $L_2$Boosting]
\begin{align*} (4') \quad \hat{y}^{m+1} \equiv f^{m+1} = P_m y \end{align*} 
and
\begin{align*}  U_i^{m+1}=Y_i-\hat{Y}_i^{m+1},\end{align*}
where $P_m$ denotes the projection of the variable $y$ on the space spanned by the first $m$ selected variables (the corresponding regression coefficient is denoted $\beta^{m}_o$.)
\end{algorithm}

Define $X_o^m$ as the matrix which consists only of the columns which correspond to the variables selected in the first $m$ steps, i.e. all $X_{j_k}$, $k=0,1,\ldots,m$.
Then we have:
\begin{eqnarray}
 \beta^{m}_o &=& ({X_o^m}' X_o^m)^{-1} {X_o^m}' y\\
\hat{y}^{m+1}=f_o^{m+1}&=& X_o^m \beta^{m}_o
\end{eqnarray}

\section{Inference with $L_2$Boosting}
In many cases the researcher is interested in valid inference of a low-dimensional parameter $\alpha$ in the presence of a high-dimensional nuisance parameter $\eta$ where modern methods of machine learning are used to estimate the nuisance parameter $\eta$. To achieve this, two conditions are sufficient. First, it is important that estimating equations used to draw inferences about $\alpha$ satisfy a key orthogonality or immunization condition. When estimation and inference for $\alpha$ are based on the system of equations $M(\alpha, \eta)=0$, this condition is fulfilled if
$$ \partial_{\eta} M(\alpha, \eta)=0.$$ This is an important element in providing an inferential procedure for $\alpha$ that remains valid when $\eta$ is estimated using regularized machine learning methods, like Lasso or boosting. This orthogonality condition can generally be established. Second, it is important to use high-quality, structured estimators of $\eta$. Additional structure is usually required for $\eta$ and the imposed estimator shall replicate this structure. It has been shown that under a sparsity condition Lasso is such a high-quality estimator. For a detailed description of this approach to valid post selection inference we refer to another article in this session, \cite{CCDDHN:2016}, and \cite{CHS:2016}.

\cite{SL_2016} show that both post and orthogonal $L_2$Boosting fulfill the conditions for a high quality estimator and have the same rate of convergence as Lasso.\footnote{They also derive an upper bound for the rate of convergence of $L_2$Boosting which is slower than the Lasso rate. Under additional assumptions an improvement might be possible.} Hence, those two variants of boosting can be used in a high-dimensional setting to estimate the nuisance parameter $\eta$ and finally provide valid inference for the target parameter of interest, $\alpha$. Formal statements of those results are provided in \cite{SL_2017}. In the following sections we show how $L_2$Boosting can be used in practical economic problems, namely IV estimation with many potential IVs, and inference on treatment effects after selection among high-dimensional controls with $L_2$Boosting. \cite{BCCH12} and \cite{BelloniChernozhukovHansen2011} provide the underlying theory for high-dimension, including the immunized moment conditions.

\section{Estimation of Treatment Effects}
In this section we report briefly the results from the simulation studies and the applications. A detailed description can be found in the supplement to the paper.

\subsection{IV estimation with many instruments}
Here we use boosting for estimation of the first stage in an IV setting with very many potential instrumental variables. The simulations show that boosting performs well in common settings, giving a lower bias than post-Lasso and rejection rates close to the nominal $5\%$ level. In the application the influence of property rights protection, measured by federal appellate court decisions regarding in eminent domain, on GDP is analyzed. The boosting estimates replicate the Lasso estimates but with smaller standard errors. The economic conclusions remain unchanged.

\subsection{Inference on treatment effects after selection among high-dimensional controls}
For estimation and inference of a treatment effect in a setting with very many control variables we apply the so-called double selection method. It consists of two separate selection steps to determine the final controls for the regression of the outcome variable on the treatment variable and the selected controls. In the simulations Lasso shows a lower bias, but the rejection rates are too small compared to the nominal $5\%$ level. The boosting estimates are close to the nominal level. The application, analyzing the convergence hypothesis in Macroeconomics, replicates the Lasso estimates.

\section{Conclusion}
In this paper we define and explain briefly $L_2$Boosting and two variants, namely post-$L_2$Boosting and orthogonal boosting. We show, how these methods can be used within the orthongal moment conditions framework for valid post-selection inference on treatment effects, either in an IV estimation setting with many instruments or in a setting with very many controls. Although only post- and orthogonal boosting have been shown to have rate of convergence in prediction norm, allowing valid post-selection inference, all three versions show very good properties in the simulations. In the applications we present boosting replicates the results from Lasso estimates. In sum, boosting seems to be an useful tool for the (micro-)econometrician's toolbox. The strength of boosting is particular in more complex models and offers many interesting questions for future research.

\bibliographystyle{aea}
\bibliography{Literatur_NR, mybibAR}

\end{document}


\begin{center}
{\huge \textsf{Online Appendix for ``$L_2$Boosting for Economic Applications''}}\\
\bigskip

\end{center}

\title[$L_2$Boosting for Economic Applications]{}
\author{Ye Luo and Martin Spindler}\thanks{Luo: University of Florida.
Spindler:  University of Hamburg, Hamburg Business School, Moorweidenstr. 18, 20148 Hamburg, Germany, martin.spindler@uni-hamburg.de. }
\date{December, 2016. We thank seminar participants and the discussant Hai Wang at
the AEA Session on Machine Learning in Econometrics for useful comments.}
\maketitle

\begin{footnotesize}
\textbf{Abstract.} In this supplement additional material, in particular simulation results and applications, for the paper \textquotedblleft $L_2$Boosting for Economic Applications\textquotedblleft is presented.

\textbf{Key words:} $L_2$Boosting, High-dimensional, instrumental variables, treatment effects, post-selection inference.

\end{footnotesize}

In this supplement additional material for the article \textquotedblleft $L_2$Boosting for Economic Applications\textquotedblleft is presented. First, a brief literature review of using boosting in Economics and Finance is given. The main part shows -- by simulations and applications -- how boosting can be used for estimation of treatment effects in a setting with very many control variables and with very many potential instrumental variables. 


\section{A Brief Review of the Literature}
In this section we give a very brief review of applications of boosting in Economics and Finance. As the strength of machine learning is in prediction and model selection, boosting has been mainly used in these domains. Although boosting has been shown to be a useful approach in many statistical applications, it has been more or less ignored in empirical economics and finance. Some of the few exceptions include the following applications. 

Boosting has been used for modeling and predicting volatility,amongst others, by \cite{MRS_2015}, \cite{nr:audrino.buhlmann:2003} and \cite{nr:audrino.buhlmann:2009}. \cite{nr:audrino.buhlmann:2009} model stock--index volatility in a GARCH framework and employ boosting for componentwise knot selection in bivariate--spline estimation. \cite{MRS_2015} also employ a GARCH framework for modeling volatility but allow for a large set of macroeconomic variables which drive volatility. Their data set consists of monthly data with 253 months in total and 40 macroeconomic variables leading to more than 80 predictors (allowing lags). They employ boosting for model estimation and variable selection.

\cite{nr:bai.ng:2009} use boosting to select the predictors in factor-augmented autoregressions. \cite{Ng:2013} classifies and predicts recessions with boosting.


\section{IV estimation with many instruments}
In this section we demonstrate how boosting can be used for IV estimation in a setting with very many instruments.

\subsection{Simulation}
The simulations are based on a simple instrumental variables model data-generating process (DGP):
\begin{align}
y_i &= \beta d_i + e_i,\\
d_i &= z_i \Pi + v_i,\\
(e_i,v_i) &\sim N \left(0 ,\left( \begin{array}{cc}
	\sigma_e^2 & \sigma_{ev}\\
	\sigma_{ev} & \sigma_v^2
\end{array} \right) \right) i.i.d.,
\end{align}
where $\beta=1$ is the parameter of interest. The regressors $Z_i=(z_{i1}, \ldots, z_{i100})'$ are normally distributed $N(0, \Sigma_Z)$ with $\mathbb{E}[z_{ih}^2]=\sigma^2_z$ and $Corr(z_{ih}, z_{ij})=0.5^{|j-h|}$. $\sigma^2_z$ and $\sigma^2_e$ are set to unit, $Corr(e,v)=0.6$. $\sigma^2_v=1-\Pi'\Sigma_z \Sigma$ so that the the unconditional variance of the endogenous variable equals $1$. The first stage coefficients are set according to $\Pi= C \tilde{\Pi}$. For $\tilde{\Pi}$ we use a sparse design, i.e., $\tilde{\Pi}=(1, \ldots,1,0,\ldots,0)$ with $s$ coordinates equal to one and all other $p-s$ equal to zero. $C$ is set in such a way that we generate target values for the concentration parameter $\mu = \frac{n \Pi'\Sigma_z \Pi}{\sigma_v^2}$ which determines the behavior of IV estimators.
We set the sample size equal $n$ to $100$, $s=5$, $p=100$ and the concentration parameter equal to $180$.  We estimate the first stage and calculate the first stage predictions with $L_2$Boosting and its variants. The simulation results in Table \ref{Sim_IV} reveal that boosting has a smaller bias than post-Lasso in this setting. While post-Lasso produces rejection rates below the nominal $5\%$ level, boosting is slightly above.

\begin{table}
\caption{Simulation results.}
\label{Sim_IV}
\begin{tabular}{lllll}
\hline \hline
&  post-Lasso & BA & post-BA & oBA \\
\hline 
bias & $0.194$ &   $0.142$&   $0.142$ &  $0.141$\\  
RP &  $0.032$ &  $0.060$&  $0.064$ &  $0.056$\\   
\hline \hline
\end{tabular}
\end{table}

\subsection{Application}
We consider IV estimation of the effects of federal appellate court decisions regards in eminent domain on macroeconomic outcomes, here in particular the log of the GDP.\footnote{We refer to \cite{BCCH12} for more information on this application.} The structural model is given by
\begin{align}
y_{ct} = \alpha_c + \alpha_t + \gamma_{c}t + \beta Takings Law_{ct}+ W_{ct}'\delta + \varepsilon_{ct},
\end{align}
where $y_{ct}$ is the economic outcome, here log of GDP, for circuit $c$ at time $t$, $Takings Law_{ct}$ number of pro-plaintiff appellate takings decisions in circuit $c$ and time $t$, $W_{ct}$ judicial pool characteristics, a dummy for whether there were no cases in that circuit-year, and the number of takings appellate decisions; $\alpha_c$, $\alpha_t$ and $\gamma_c t$ denote circuit-specific, time-specific and circuit-specific time trends. The parameter of interest, $\beta$, represents the effect of an additional decision upholding individual property interpreted as more protective individual property rights. The sample size is $312$. The analysis of the causal effect of takings law is complicated by potential endogeneity between taking law decisions and economic variables. We employ an instrumental variables strategy that relies on the random assignment of judges to federal appellate panels and uses characteristics of federal circuit court judges (e.g. gender ,race, religion, political affiliation, etc.) as instruments. This gives $138$ instruments. We estimate the effect $\beta$ by doing the selection of IVs and estimation the first-stage predicted values $\hat{Takings Law}_{ct}$ by employing the boosting algorithms introduced before. The results are given in Table \ref{Appl_IV}. The boosting estimates agree with the Lasso estimate but give smaller standard errors. The economic conclusions remain unchanged.

\begin{table}
\caption{Effect of Federal Appellate Takings Law Decisions on Economic Outcomes.}
\label{Appl_IV}
\begin{tabular}{lllll}
\hline \hline
&  post-Lasso & BA & post-BA & oBA \\
\hline 
$\hat{\beta}$ & $0.005$ &   $0.005$&   $0.004$ &  $0.008$\\ 
se &  $0.012$ &  $0.007$&  $0.006$ &  $0.006$\\
\hline \hline
\end{tabular}
\end{table}

\section{Inference on treatment effects after selection among high-dimensional controls}
\subsection{Simulation}
Here we consider the following data-generating process:
\begin{align}
y_i &= d_i \alpha_0 + x_i' \theta_g + \xi_i\\
d_i &= x_i' \theta_m + \nu_i,
\end{align}
where $(\xi_i, \nu_i)' \sim N(0,I_2)$ with $I_2$ the $2 \times 2$ identity matrix, $p=200$, $x_i \sim N(0, \Sigma)$ with $\Sigma_{kj}0=.5^{|j-k|}$. The parameter of interest, $\alpha_0$, is set equal to $0.5$ and the sample size is $n=100$. We consider a design with a decaying sequence of $\theta_m$ and $\theta_g$, namely $1/j^2$ for $j=1,\ldots,p$. The results in Table \ref{Sim_TE} show that post-Lasso has a smaller bias than the boosting algorithms, but too small rejection rates (RP) compared to the nominal $5\%$ level. Boosting has rejection rates close to the nominal level.

\begin{table}
\caption{Simulation results.}
\label{Sim_TE}
\begin{tabular}{lllll}
\hline \hline
&  post-Lasso & BA & post-BA & oBA \\
\hline 
bias & $0.082$ &   $0.121$&   $0.136$ &  $0.121$\\
RP &  $0.002$ &  $0.042$&  $0.054$ &  $0.042$\\ 
\hline \hline
\end{tabular}
\end{table}

\subsection{Application}
In Macroeconomics an important questions is how the rates ($Y$) at which economies of different countries grow are related to the initial wealth levels in each country ($D$) controlling for country's institutional, educational, and other similar characteristics ($W$). The relationship is captured by $beta_1$, the ``speed of convergence/divergence'', it measures the speed at which poor
  countries catch up $beta_1< 0$ or fall behind $beta_1> 0$ rich countries, after controlling for $W$. Hence the model is given as 
	\begin{align} Y &= \beta_1 D + \beta_2' W + \epsilon. \end{align}
	For the analysis we use the Barro-Lee data set with $90$ countries (observations) and about $60$ controls. We estimate the parameter of interest by the double selection method employing both Lasso and $L_2$Boosting for the two selection steps. The double selection method implicitly constructs an orthogonal moment condition. The results are given in Table \ref{growth}. Here again, the boosting estimates agree with the Lasso estimate and confirm the convergence hypothesis.
	
\begin{table}[h!]
\caption{Effect of Initial GDP level om Growth.}
\label{growth}
\begin{tabular}{lllll}
\hline \hline
&  post-Lasso & BA & post-BA & oBA \\
\hline 
$\hat{\beta}$ & $-0.040$ &  $-0.042$& $-0.042$ & $-0.041$\\ 
se & $0.015$ & $0.012$& $0.012$ & $0.013$\\
\hline \hline
\end{tabular}
\end{table}
\bibliographystyle{aea}
\bibliography{Literatur_NR}